\documentclass[conference]{IEEEtran}

\ifCLASSINFOpdf
\else
\fi

\usepackage[tight,footnotesize]{subfigure}

\usepackage{graphicx}

\usepackage{caption}
\usepackage[font=footnotesize]{subfig}

\DeclareRobustCommand{\IEEEauthorrefmark}[1]{\smash{\textsuperscript{\footnotesize #1}}}

\begin{document}

\title{A Data Cube of Big Satellite Image Time-Series for Agriculture Monitoring}

\author{\IEEEauthorblockN{Thanassis Drivas$^*$\IEEEauthorrefmark{1}, Vasileios Sitokonstantinou$^*$\IEEEauthorrefmark{1,2}, Iason Tsardanidis\IEEEauthorrefmark{1},\\ Alkiviadis Koukos\IEEEauthorrefmark{1}, Charalampos Kontoes\IEEEauthorrefmark{1} and Vassilia Karathanassi\IEEEauthorrefmark{2}}
\\
\IEEEauthorblockA{\IEEEauthorrefmark{1}National Observatory of Athens, IAASARS, BEYOND Center, Penteli, Greece\\
Email: \{tdrivas, vsito, j.tsardanidis, akoukos, kontoes\}@noa.gr}

\IEEEauthorblockA{\IEEEauthorrefmark{2}Laboratory of Remote Sensing, National Technical University of Athens, Athens, Greece\\
Email: karathan@survey.ntua.gr
}
}

\maketitle
\def\thefootnote{*}\footnotetext{These authors contributed equally to this work}

\begin{abstract}
The modernization of the Common Agricultural Policy (CAP) requires the large scale and frequent monitoring of agricultural land. Towards this direction, the free and open satellite data (i.e., Sentinel missions) have been extensively used as the sources for the required high spatial and temporal resolution Earth observations. Nevertheless, monitoring the CAP at large scales constitutes a big data problem and puts a strain on CAP paying agencies that need to adapt fast in terms of infrastructure and know-how. Hence, there is a need for efficient and easy-to-use tools for the acquisition, storage, processing and exploitation of big satellite data. In this work, we present the Agriculture monitoring Data Cube (ADC), which is an automated, modular, end-to-end framework for discovering, pre-processing and indexing optical and Synthetic Aperture Radar (SAR) images into a multidimensional cube. We also offer a set of powerful tools on top of the ADC, including i) the generation of analysis-ready feature spaces of big satellite data to feed downstream machine learning tasks and ii) the support of Satellite Image Time-Series (SITS) analysis via services pertinent to the monitoring of the CAP (e.g., detecting trends and events, monitoring the growth status etc.). The knowledge extracted from the SITS analyses and the machine learning tasks returns to the data cube, building scalable country-specific knowledge bases that can efficiently answer complex and multi-faceted geospatial queries.

\end{abstract}


\IEEEpeerreviewmaketitle


\section{Introduction}
The Common Agricultural Policy (CAP) has set out to implement radical changes towards fairer, greener and more performance-based policies \cite{newcap}. In this context, and inspired by the advent of free and open satellite data and the recent advancements in data science, the CAP aims at the country-wide evidence-based  monitoring of the farmers' compliance with the agricultural policies. Towards this effort, the Sentinel satellite missions, advanced ICT technologies and Artificial Intelligence (AI) have been identified as key enablers \cite{datacap}. 

The Sentinels provide frequent optical and Synthetic Aperture Radar (SAR) images of high spatial resolution and have been extensively used for the monitoring agriculture and specifically for the purposes of the CAP \cite{cap2, semantics}. Most Sentinel-based CAP monitoring systems utilize the parcel boundaries from the Land Parcel Identification System (LPIS). LPIS is a database that connects the crop type label, as declared by the farmer, to each parcel object. The LPIS objects and crop labels are then combined with the Satellite Image Time-Series (SITS) to feed AI models for CAP monitoring applications \cite{datacap}.

CAP monitoring systems need to be able to process and visualize  large amounts of satellite data, which is not possible using traditional local storage and processing workflows, and for this reason big Earth Observation (EO) management technologies are necessitated \cite{cubes_lewis}. One such example are the EO Data Cubes (EODCs) that can handle large volumes and provide a solid solution for accessing and managing Analysis Ready Data (ARD) \cite{cubes_giuliani}. Currently, several EODC technologies have been developed, e.g., Google Earth Engine \cite{gee_official}, Sentinel-hub \cite{sentinelhub}, gdalcube \cite{gdalcube}, Rasdaman \cite{RasDaMan_official}, Open Data Cube (ODC), OpenEO \cite{openeo}, and Earth System \cite{earth_system}.

In this work, we demonstrate an end-to-end workflow for building and exploiting a scalable Agriculture monitoring Data Cube (ADC) based on ODC, which is hosted on CreoDIAS, one of the five Data and Information Access Services (DIAS) cloud platforms. ODC is open and infrastructure-independent, so it can be installed in diverse environments, from personal computers to supercomputers \cite{odc_countries}. It enables effortless data management and simplifies data querying, using a Python Application Programming Interface (API) \cite{cubes_catalan}. In ODC-based systems, data are indexed or ingested into data cubes and can be then loaded into xarrays, which is a  powerful multidimensional data structure. The simplification of processing and exploitation of big spatio-temporal data, using data cubes, allows for unlocking their full potential and strengthens the connection between data and users \cite{cubes_kopp}. The ODC technology has been used for the implementation of a number of national data cubes, i.e., Australia \cite{odc_australia}, Africa \cite{odc_africa}, Switzerland \cite{odc_swiss}, Colombia \cite{odc_columbia}, Brazil \cite{odc_brazil} and Taiwan \cite{odc_taiwan}.

Our solution comes with several notable advantages when compared with the related work, as it includes both satellite data (SAR and optical) and auxiliary geospatial data, such as the farmers' declarations (crop type labels) and parcel boundaries from the LPIS, opening the door to fast and accurate operations between them. This is crucial for operational scenarios of CAP monitoring applications. In addition, we build several tools on top of the ADC. Specifically, we built functionalities that enable the effortless generation of corrected, cleaned and smoothed SITS that are formed into feature spaces that feed AI pipelines with ARD. We also offer functionalities for computing multidimensional statistics and in turn smart geospatial queries, enabling the recognition of patterns and trends and the detection of events on agricultural land. The outputs of the AI models (e.g., crop classification and grassland mowing detection) fed by ADC, along with the multidimensional information extracted from the monitoring of an area throughout the years, populate the cubes through a feedback loop. This way, we generate unique country-specific EO knowledge bases for agriculture monitoring. These knowledge bases allow for the evidence-based decision-making by fully exploiting big EO data products and results of AI models.


\section{Agriculture monitoring data cube}\label{sec:ii}
The architecture of the ADC is based on two layers. The first layer is related to data discovery and acquisition, and the second layer is for the production of ARD. The ARD generation layer includes a number of pre-processing tools for SAR and optical images that make use of open-source libraries. Instructions and code for setting up a data cube like ADC can be found in https://github.com/Agri-Hub/datacap. 

\subsection{Access to Sentinel data}
Copernicus Sentinel missions and specifically Sentinel-1 (SAR) and Sentinel-2 (optical) have contributed to the monitoring of agriculture by providing high temporal and high spatial resolution images at global scale. Sentinel-2 offers images in the visible and near infrared parts of the spectrum, making them ideal for vegetation monitoring. Apart from the surface reflectances of Sentinel-2, vegetation indices, e.g., the Normalized Difference Vegetation Index (NDVI), can also be extracted from optical images to enhance certain vegetation characteristics (e.g., water content, physiological stress etc.). SAR images (Sentinel-1) have also been used in related work, either as stand-alone or complementary to optical images, as they are not affected by clouds and allow for constructing dense SITS that are essential in agriculture monitoring \cite{s1cap2}. 

To generate ARD we need to consider both data acquisition and data storage. To minimize the effort of accessing data, ADC has been developed within the CreoDIAS cloud platform. This allows us to access data directly via an offered catalogue (eodata), which according to the latest statistics includes more than 27 PB of Sentinel-1 and Sentinel-2 data \cite{creodias}. The CreoDIAS object storage repository ensures good performance and eliminates the need for repeated downloading of raw data locally. It is worth mentioning that CreoDIAS hosts the complete archive of Sentinel data, and thus we do not have to search across multiple data hubs. This is often a complicated and strenuous process due to the large number of sources and their varying performances (i.e., download speed, latency etc.).

Our ADC includes both Sentinel-1 and Sentinel-2 products that cover Cyprus and Lithuania over the span of three years (2019-2021). This results to a total number of approximately 15 TB of data, which are pre-processed before indexed in the ADC. The metadata of the products, as retrieved from the CreoDIAS API, are stored to a database. Thus, spatial operations are allowed giving the potential for statistics extraction. In addition, flags per product are created in every step of the pre-processing chain aiming at the monitoring of possible failures and the re-execution of the problematic task, if needed.

\subsection{Generation of Analysis Ready Data}
In order to convert the raw satellite data to ARD, certain processing steps are required, i.e., i) atmospheric corrections for optical images (Sentinel-2) using the Sen2Cor software (Level 2A products) and ii) the generation of backscatter coefficients ($\sigma_{0}$) and interferometric coherence for the SAR data (Sentinel-1) using the python library snappy. Then all the products were converted to Cloud Optimized TIFFs that are more efficient in terms of storage and loading speed and thus fast enough for web applications. In addition, as ADC handles data from two sensors, it is important for all the pixels to have the same spatial resolution and be spatially aligned. For this, Sentinel-1 and Sentinel-2 data are resampled to 10 m and matched pixel-to-pixel, thus enabling efficient, sensor-agnostic time-series data analysis. The final step of the pre-processing chain is the masking of clouds and shadows on the optical images; the detection of which is done using the Sentinel-2 scene classification product of the Sen2Cor algorithm.

 
 
\begin{figure}[h!]
\centering
\includegraphics[scale=1.4]{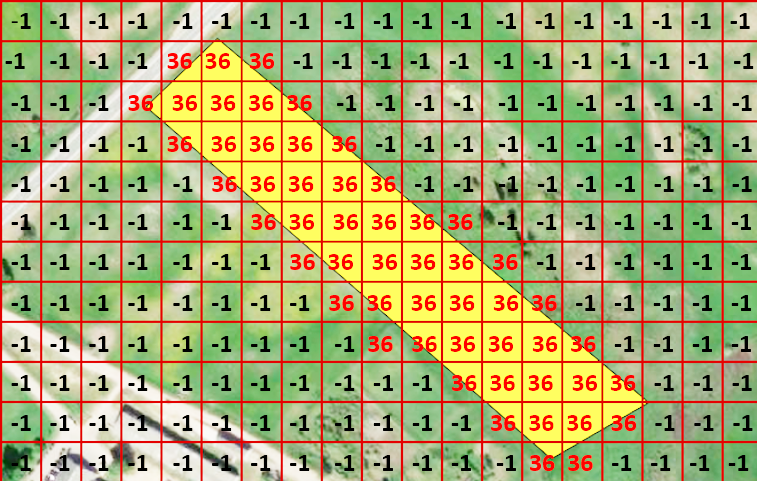}
\caption{Example of how parcels are indexed in the data cube as rasters. Value $-1$ represents the pixels that do not belong to any stored parcel, whereas pixels intersecting a parcel are labeled with it's corresponding database id (here $36$).}
\label{fig:geom_index}
\end{figure}

ADC is able to scale efficiently and cover large areas, which possibly include millions of objects (parcels), for which we need to compute statistics and execute geospatial queries. Besides serially querying the data cube for each parcel, using its boundaries in vector format, we can also rasterize them as shown in Fig.\ref{fig:geom_index}. Then we index the layer of rasterized parcels and load them into the data cube. This enables the fast and parallel computation of zonal statistics per parcel using the powerful xarrays data structure. We use the \textit{groupby} function, which allows for the grouping of the xarray dimensions based on the respective IDs. Table \ref{tab:processing_time} shows the processing time for generating zonal statistics for a varying number of parcels using the serial querying method and the groupby method. It is observed that for a large number of parcels, groupby is far more computationally economic. 



\begin{table}[!ht]
\caption{Execution time for generating monthly averages for one year and one Sentinel-2 band over one tile.}
\label{tab:processing_time}
\centering
\scalebox{1.3}{
\begin{tabular}{|c|c|c|}
\hline
\textbf{\# parcels}& 
\textbf{groupby} & \textbf{serial querying}\\ \hline
$1$ k & $69$ sec & $250$ sec\\ \hline
$10$ k & $71$ sec & $40$ min\\ \hline
$100$ k  & $150$ sec & $400$ min\\ \hline
\end{tabular}}
\end{table}


\section{Satellite image time-series analysis}
Long and dense SITS are essential for agriculture monitoring since crops change dynamically with time. We need to capture these changes to i) detect trends and events (e.g., grassland mowing detection), ii) monitoring the growth status and health, iii) classify crops to verify the validity of declarations and many more applications pertinent to the monitoring of the CAP. We enable the effortless, rapid and large scale analysis of long and dense SITS, by exploiting the capabilities of xarrays. In this section, practical scenarios are described to showcase in detail the functionalities that are built on top of the ADC and specifically how we can reduce noise, detect trends, speed up work and improve reliability of decision-making. The code and data are available at https://github.com/Agri-Hub/ADC. 

\subsection{Spatial buffers}
\begin{figure}[h!]
\centering
\includegraphics[scale=0.3]{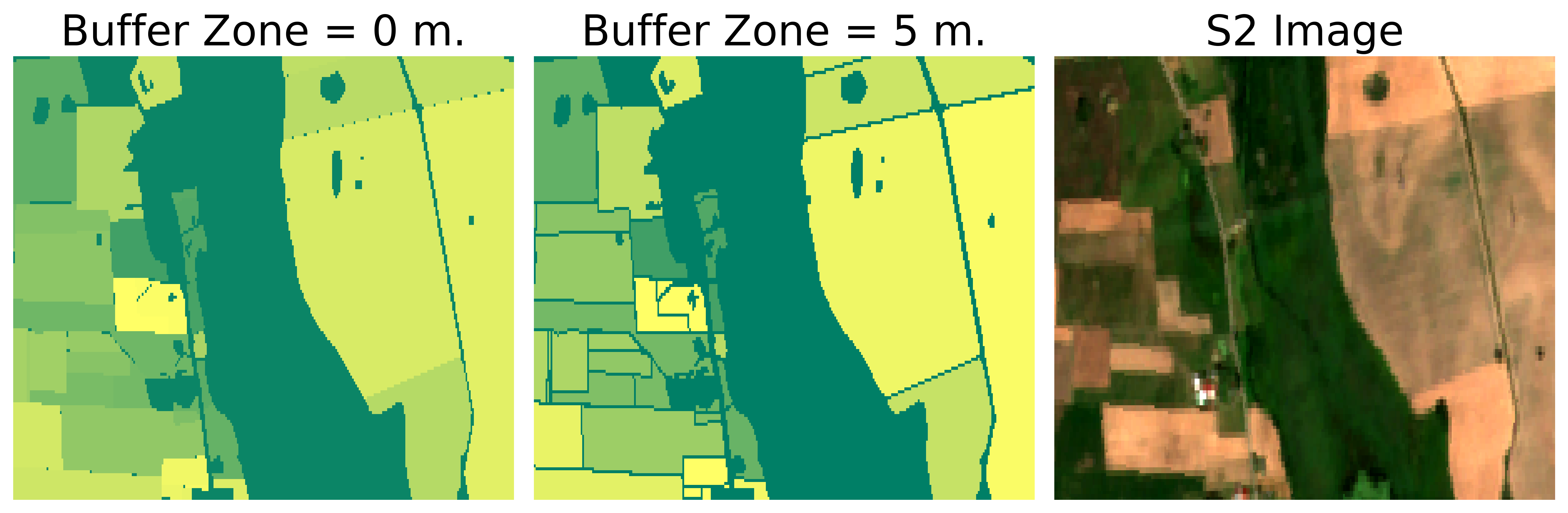}
\caption{Inward buffer to parcel boundaries to avoid using mixed pixels.}
\label{fig:parcels_buffering}
\end{figure}

One of the main advantages of ADC is the rasterized information of the parcels, described in Sec. \ref{sec:ii}. Apart from the real parcel geometries, one can use our inward buffer functionality, as shown in Fig.\ref{fig:parcels_buffering}, which can alleviate the adverse consequences of mixed pixels. This way, we can extract only the representative information from the rest of the pixels encompassed inside the parcels' buffered geometries.

\begin{figure}[ht!]
\centering
\includegraphics[scale=0.3]{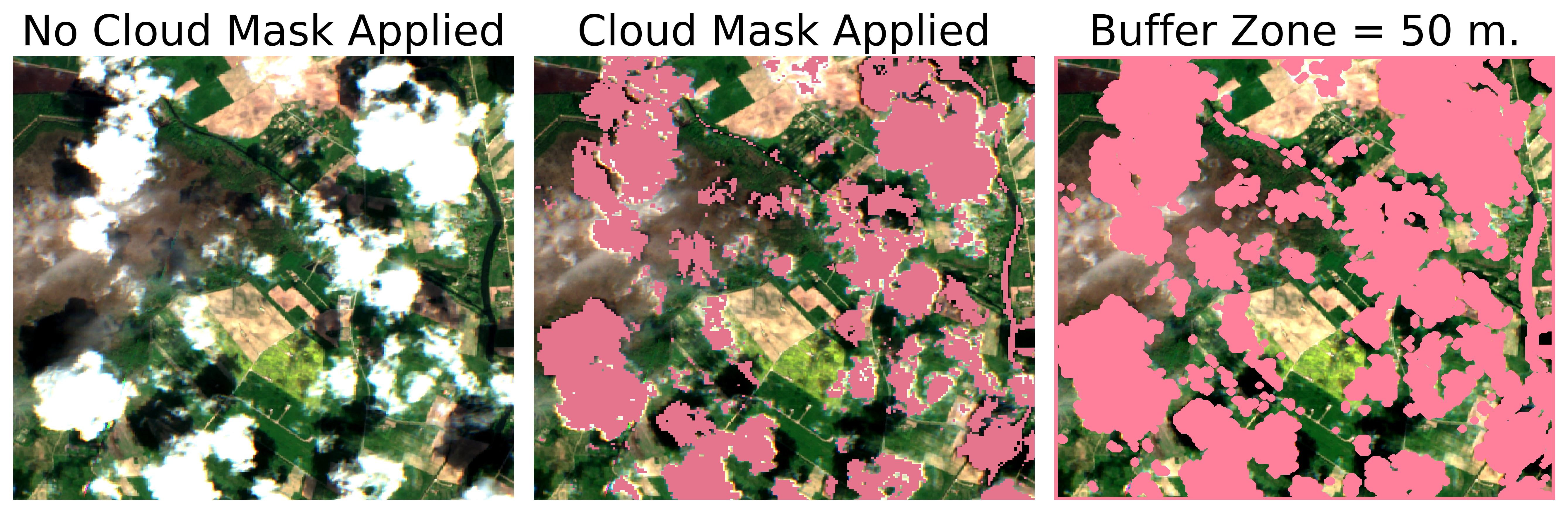}
\caption{Applying an outward buffer on cloud and shadow objects.}
\label{fig:cloud_masking}
\end{figure}

We additionally provide an adjustable outward buffer that can be applied on the less-than-perfect cloud and shadow mask products in order to reduce some of the noise. The Sen2Cor scene classification product has suboptimal recall for the cloud and shadow classes. However, for many agriculture monitoring applications it is important to have only clear pixels involved in the analyses. To tackle this issue, one can apply an outward buffer zone around the masked cloud and shadow objects. As a result, the pixels adjacent to clouds are now classified as cloudy, providing a trade-off between better cloud masking and fewer clear pixels for analysis. The original cloud mask and a cloud mask with a buffer are illustrated in Fig. \ref{fig:cloud_masking}.

\subsection{SITS preparation}

\begin{figure}[h!]
\centering
\includegraphics[scale=0.25]{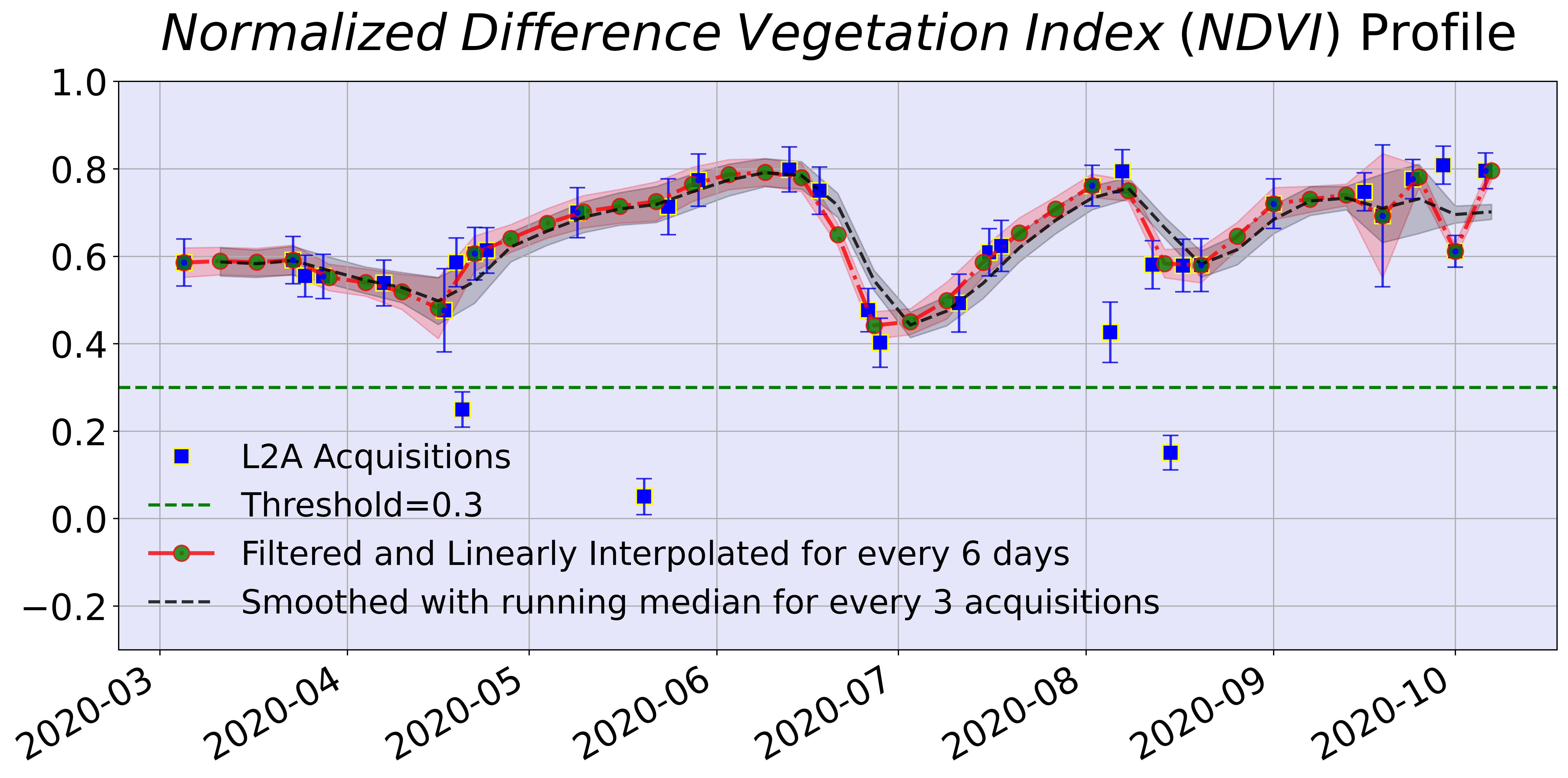}
\caption{An example of SITS preparation functionalities (filtering, interpolation, resampling and smoothing) applied to all pixels of a sampled parcel. }.
\label{fig:ts_preprocessing}
\end{figure}

Our ADC solution offers a number of SITS processing techniques to filter, interpolate, resample and smooth the pixel or parcel time-series. For instance, by applying specific thresholds we can filter out outliers, as seen in Fig. \ref{fig:ts_preprocessing}, and fill the missing values from clouds and shadows using a number of off-the-shelf interpolation techniques that we offer (e.g., Linear, Bicubic etc.). Interpolated SITS can then be sampled at any desired temporal resolution. This is useful when constructing feature spaces for machine learning tasks that require homogeneous input of fixed elements. Additionally, our solution provides smoothing functionalities (e.g., rolling median) to eliminate the noise caused by temporal fluctuations on the SITS. Thus, patterns can be more clear and reveal trends throughout the year(s). By refining the SITS, one can enhance the performance of AI models that are fed with these data, as well as improve photo-interpretation tasks. For example, crop classification tasks perform better using interpolated time-series and grassland mowing event detection is significantly enhanced through filtering and smoothing that removes outliers and abrupt changes that could be mistaken for real events.

\subsection{Feature space generation}\label{sec:fs_generation}
Using the ADC, one can go beyond single-date image features and combine spectral bands and vegetation indices that come from multiple temporal instances and thereby extract phenological features via computing the integrals and derivatives of the time curves (i.e., biomass indicator, yield indicator, end-of-season, peak-of-season, start-of-season) \cite{sitokonstantinou2020sentinel}. Such features have been shown to enhance the performance of crop classification AI tasks \cite{phenology_ids_crop_classification_Htitiou_1}. The capacity of the ADC to navigate in the spatial, temporal and spectral/variable dimensions, using the grid format of multidimensional xarrays, enables the effortless generation of SITS feature spaces. One can select to generate a feature space at the pixel or parcel level to match the downstream application requirements for spatial resolution. Additionally, one can segment the satellite images to chunks and generate patch-based datasets to feed deep learning models for computer vision tasks.

\subsection{Smart multidimensional queries}

Time-series of earth observations hold important information on the evolution of crops and should be analyzed to extract knowledge on the vegetation development and to identify potential trends. By using the ADC, one can easily perform historical statistical analysis over an area at different time units, i.e., day, month, season or year. These statistics can be provided in the form of aggregated values for a parcel or any user-defined area. Apart from the coordinates of the area, a user can decide on additional parameters, such as the maximum cloud coverage percentage or the minimum number of cultivated crop fields over this area etc. 

\begin{figure}[h!]
\centering
\includegraphics[scale=0.3]{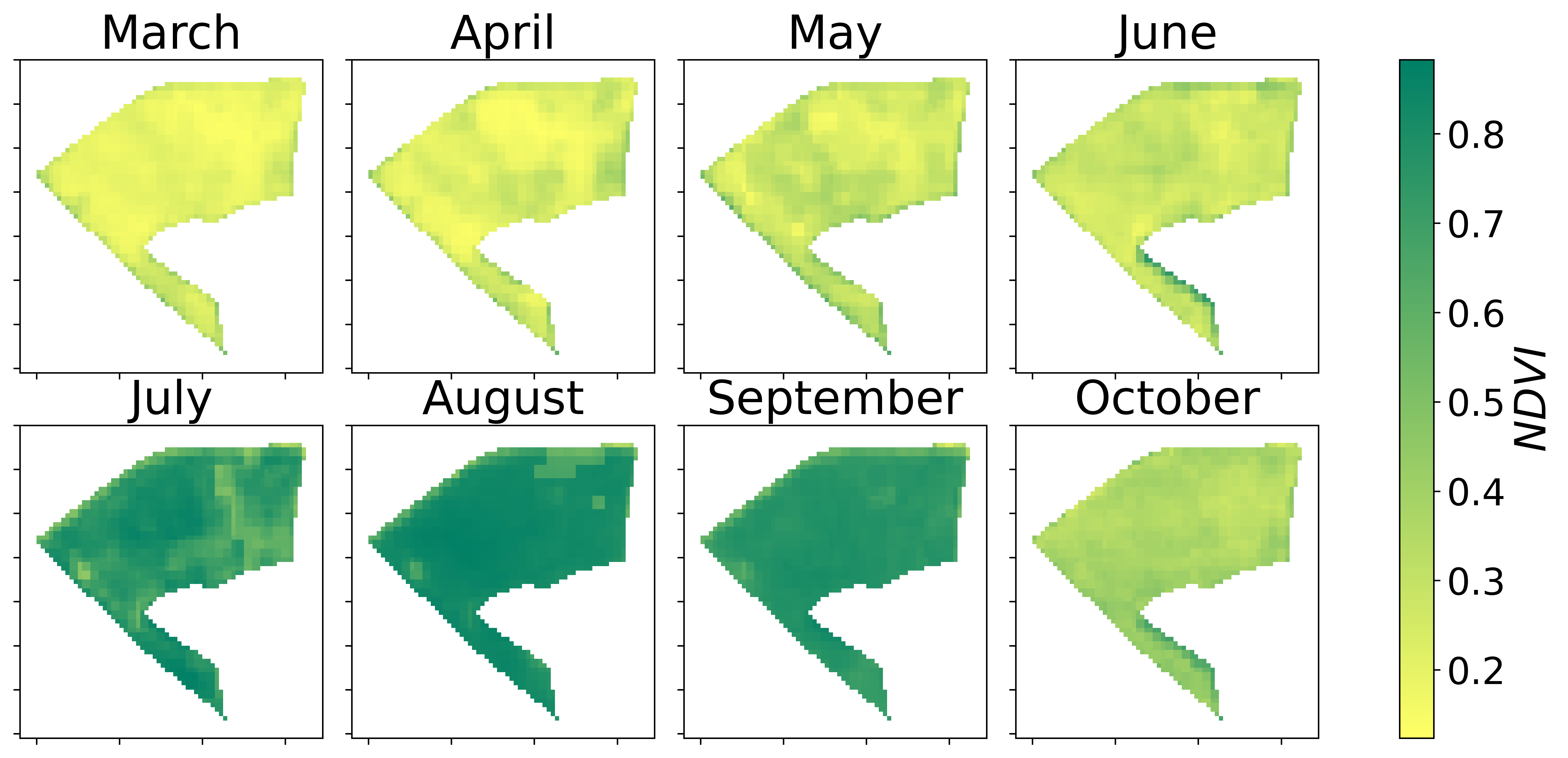}
\caption{An example of NDVI time-series animation for a maize field with monthly averages.}
\label{fig:analytic1}
\end{figure}

Using the ADC, we can generate animations of the evolution of a selected Sentinel variable (e.g., NDVI). Fig. \ref{fig:analytic1} shows a useful example on how temporal statistics can be used for validating if a CAP obligation is met or not. Specifically, the illustrated field was declared as \textit{spring triticale}, which is a particular type of \textit{spring cereal}, and it was predicted, by our AI crop classifier \cite{semantics} to be \textit{maize}, which is a summer cultivation. The animation in Fig. \ref{fig:analytic1} can be used to verify that indeed the prediction is correct and the declaration is erroneous, since the field has substantial vegetation during the summer months and spring triticale would have been harvested. This functionality is particularly useful for paying agency inspectors of the CAP that are not EO experts. This way, they can easily fetch a parcel-focused Sentinel time-series for the field of inspection and decide on the validity of the farmer's declaration.

Similarly to the temporal statistics described above, one can use ADC to compute pertinent statistics along the space and spectral/variable dimensions. All these statistics can be used in a synergistic manner to formulate complex geospatial queries. This is a game changer for the inspectors of the CAP paying agencies that are for the first time enabled to ask combinatory and multidimensional questions to make their inspection process faster, targeted and precise. As already mentioned, the ADC supports several services (e.g., crop classification, grassland mowing detection, SITS multidimensional statistics etc.) by providing gridded ARD. The results of these services are used to update a PostgreSQL/PostGIS database. The database contains aggregated results per parcel, which can then be easily accessed, enabling a back-and-forth communication with the ADC.

Thus, we have data cubes that include and keep on being dynamically populated by Sentinel-1 and Sentinel-2 products.  We also have auxiliary geospatial data (e.g., LPIS) that are used to enable the provision high level data products (e.g., crop classification) that in turn populate the cubes. This way, we end up with country specific knowledge bases for CAP monitoring. Meanwhile, useful operations such as the computation of distance between two geometries, the calculation of an area, buffer analysis and geospatial queries, can take place exploiting the power of the PostGIS extension. PostGIS and ADC queries can be combined to address the most sophisticated of questions inspectors might have. Below are listed examples of smart queries supported by our ADC-based framework.
\\
\textbf{Query 1.} \textit{Generate feature space}. Return monthly averages of Sentinel-1 coherence and Sentinel-2 NDVI at the parcel-level for Lithuania from 2017 to 2021 and Paphos, Cyprus for 2020. Apply inward buffer 5 m to avoid mixed pixels and outward buffer 50 m to reduce noise from clouds. Feed data to grassland mowing detection and crop classification algorithms.\\
\textbf{Query 2.} \textit{Select potential wrong declarations}. Using the output of query 1, return the fields in Paphos in 2020 that were declared to cultivate maize but have been predicted in a different crop type class. \textit{Visual verification.} Return animation of NDVI time-series, with a 10-day step, from June to October.\\
\textbf{Query 3.} \textit{Quantify grassland use intensity}. Using the output of query 1, return the number of mowing events in Lithunia per year from 2017 to 2021. Limit the results to grassland fields with average NDVI of less than $0.4$. Identify hotspots of  low grassland intensity with an average mowing event of less than 1 per year, over the years of inspection. Ultimately, this query will enable the decision makers to suggest spatially tailored mitigation or adaptation measures for the hotspots.


\section{Conclusions}
Data cubes enable the transformation of EO data into i) analysis-ready information, ii) high-level knowledge and iii) intuitive visualizations to support timely and effective decision making. Our cloud-based approach allows for the efficient and automated discovery, pre-processing, data cube indexing and analysis of big satellite data. Currently, ADC is populated with Sentinel-1 and Sentinel-2 images that cover wall-to-wall Cyprus and Lithuania for three years. It also includes the parcel boundaries, crop type maps (LPIS) and ancillary data that enable the development of downstream applications for the monitoring of CAP rules. We indicatively used the outputs of an in-house crop classification model and 
Sen4CAP's\footnote{https://github.com/Sen4CAP}

grassland mowing detection model. Furthermore, a suite of tools has been built on top of the ADC. The users of our framework can straightforwardly generate spatial buffers, multidimensional statistics, animations, time-series plots and feature spaces, and execute complex multidimensional geospatial queries. We address the challenge of CAP controls by bringing together EO products, geospatial services and extracted knowledge from validated models. This solution is a stepping stone towards the modernization of the CAP and the seamless integration of big EO data in the operating models of non-expert users. 


\section*{Acknowledgment}
This work has been supported by the ENVISION and e-shape projects, funded by European Union's Horizon 2020 research and innovation programme under grant agreements No. 869366 and No. 820852, respectively.


\bibliographystyle{IEEEtran}
\bibliography{refs}

\end{document}